\documentclass[11pt,a4paper]{article}
\usepackage[hyperref]{emnlp-ijcnlp-2019}
\usepackage{times}
\usepackage{latexsym}
\usepackage{booktabs}
\usepackage{url}
\usepackage{multirow}
\usepackage{graphicx}
\aclfinalcopy 

\title{Out-of-Domain Detection for Low-Resource Text Classification Tasks}

\author{Ming Tan \thanks{\, Equal contributions from the corresponding authors: \texttt{\{mingtan,yu,wanghaoy\}@us.ibm.com}.} $^{\dagger}$ \quad \quad Yang Yu  $^{*\dagger}$ \quad \quad Haoyu Wang  $^{*\dagger}$ \quad \quad \textbf{Dakuo Wang} $^\ddagger$ \\ \textbf{Saloni Potdar} $^\dagger$ \quad \quad \textbf{Shiyu Chang} $^\ddagger$ \quad \quad \textbf{Mo Yu} $^\ddagger$ \\
$^\dagger$ IBM Watson \quad \quad $^\ddagger$ IBM Research
}

\date{}

\begin{document}
\maketitle

\begin{abstract}
Out-of-domain (OOD) detection for low-resource text classification is a realistic but understudied task. The goal is to detect the OOD cases with limited in-domain (ID) training data, since we observe that training data is often insufficient in machine learning applications. 
In this work, we propose an \emph{OOD-resistant Prototypical Network} to tackle this \textbf{zero-shot OOD detection and few-shot ID classification} task. 
Evaluation on real-world datasets show that the proposed solution outperforms state-of-the-art methods in zero-shot OOD detection task, while maintaining a competitive performance on ID classification task. 

\end{abstract}

\section{Introduction}
Text classification tasks in real-world applications often consists of 2 components- In-Doman (ID) classification and Out-of-Domain (OOD) detection components \cite{dakuowang,OOD-alexa,doc,shamekhi2018face}. ID classification refers to classifying a user's input with a label that exists in the training data, and OOD detection refers to designate a special OOD tag to the input when it does not belong to any of the labels in the ID training dataset \cite{dai2007co}. Recent state-of-the-art deep learning (DL) approaches for OOD detection and ID classification task often require massive amounts of ID or OOD labeled data \cite{OOD-alexa}. In reality, many applications have very limited ID labeled data (i.e., few-shot learning) and no OOD labeled data (i.e., zero-shot learning). 
Thus, existing methods for OOD detection do not perform well in this setting.

One such application is the intent classification for conversational AI services, such as IBM Watson Assistant\footnote{https://www.ibm.com/cloud/watson-assistant/}. 
For example, Table \ref{tab:conv_examples} shows some of the utterances a chat-bot builder provided for training. Each class may only have less than 20 training utterances, due to the high cost of manual labelling by domain experts. Meanwhile, the user also expects the service to effectively reject irrelevant queries (as shown at the bottom of Table \ref{tab:conv_examples}). The challenge of OOD detection is reflected by the undefined in-domain boundary. Although one can provide a certain amount of OOD samples to build a binary classifier for OOD detection, such samples may not efficiently reflect the infinite OOD space. Recent approaches, such as \cite{doc}, make remarkable progress on OOD detection with only ID examples. However, such condition on ID data cannot be satisfied by the few-shot scenario presented in Table \ref{tab:conv_examples}.


\begin{table}
\centering
\small
\begin{tabular}{p{3cm}|p{3.6cm}}
\toprule
Intent Label & Example \\
\toprule
Help\_List & List what you can help me with. \\\cline{2-2}
 & Watson, I need your help \\
\cline{1-2}
Schedule\_Appointment & Can you book a cleaning with my dentist for me? \\\cline{2-2}
 & Can you schedule my dentist's appointment? \\\cline{1-2}
End\_Meeting & You can end the meeting now \\\cline{2-2}
 & Meeting is over \\
 \cline{1-2}
$\cdots$ & $\cdots$ \\
\hline
\hline
{\color{blue} OOD utterances} & {\color{blue} My birthday is coming!} \\
  \cline{2-2}
  & {\color{blue} blah blah...} \\
\bottomrule
\end{tabular}
\caption{A few-shot ID training set for a conversation service for teleconference management, with OOD testing examples.}
\label{tab:conv_examples}
\end{table}

This work aims to build a model that can detect OOD inputs with limited ID data and zero OOD training data, while classifying ID inputs with a high accuracy. 
Learning similarities with the meta-learning strategy \cite{matchingNet} has been proposed to deal with the problem of limited training examples for each label (few-shot learning). In this line of work, \textit{Prototypical Networks}~\cite{prototypeNet}, which was originally introduced for few-shot image classification, has proven to be promising for few-shot ID text classification \cite{diverse_few_shot}. However the usage of prototypical network for OOD detection is unexplored in this regard.

To the best of our knowledge, this work is the first one to adopt a meta-learning strategy to train an OOD-Resistant Prototypical Network for simultaneously detecting OOD examples and classifying ID examples. The contributions of this work are two-fold: 1) Unified solution using a prototypical network model which can detect OOD instances and classify ID instances in a real-world low-resource scenario. 2) Experiments and analysis on two datasets prove that the proposed model outperforms previous work on the OOD detection task, while maintaining a state-of-the-art ID classification performance.

\section{Related Work}
\vspace{-1mm}
\paragraph{Out-of-Domain Detection}
Existing methods often formulate the OOD task as a one-class classification problem, then use appropriate methods to solve it (e.g., one-class SVM  \cite{osvm} and one-class DL-based classifiers~\cite{deepOneClass,one-class-nn}.
A group of researchers also proposed an auto-encoder-based approach and its variation to tackle OOD tasks ~\cite{ood_sent_detect, OOD-GAN}.
Recently, a few papers have investigated ID classification and OOD detection simultaneously~\cite{OOD-alexa,doc}, but they fail in a low resource setting.

\paragraph{Few-Shot Learning} 
While few-shot learning approaches may help with this low-resource setting, some recent work is promising in this regard. For example, \cite{matchingNet,LearningFF,prototypeNet} use metric learning by learning a good similarity metric between input examples;  some other methods adapt a meta-learning framework, and train the model to quickly adapt to new tasks with gradients on small samples, e.g., learning the optimization step sizes~\cite{optiModelFewShot} or model initialization \cite{Model-Agnostic}. Though most of these approaches are explored for computer vision, recent studies suggests that few-shot learning is promising in the text domain, including text classification~\cite{diverse_few_shot,jiang2018attentive}, relation extraction~\cite{han2018fewrel}, link prediction in knowledge bases~\cite{xiong2018one} and fine-grained entity typing~\cite{xiong2019imposing}, and we put it to test with the OOD detection task.

\section{Approach}
In this paper, we target solving the zero-shot OOD detection problem for a few-shot {\bf meta-test} dataset $D=(D^{train}, D^{test})$ by training a transferable prototypical network model from large-scale {\bf independent source datasets}  $\mathcal{T}= \{T_1, T_2,..., T_N\}$ for dynamic construction of the {\bf meta-train} set.  Each task $T_i$ contains labeled training examples (note that a test set is not required in meta-train).  $D$ is different from the traditional supervised close-domain classification dataset from two folds: 1) $D^{test} $ contains OOD testing examples, whereas $D^{train}$ only includes labeled examples for the target domain. 2) The training size for each label in $D^{train}$ is limited (e.g. less than 100 examples). Such limitations prevent existing methods from efficiently training a model for either ID classification or OOD detection using $D^{train}$ only. 

\begin{figure}[t!]
  \centering
  \includegraphics[scale=0.24]{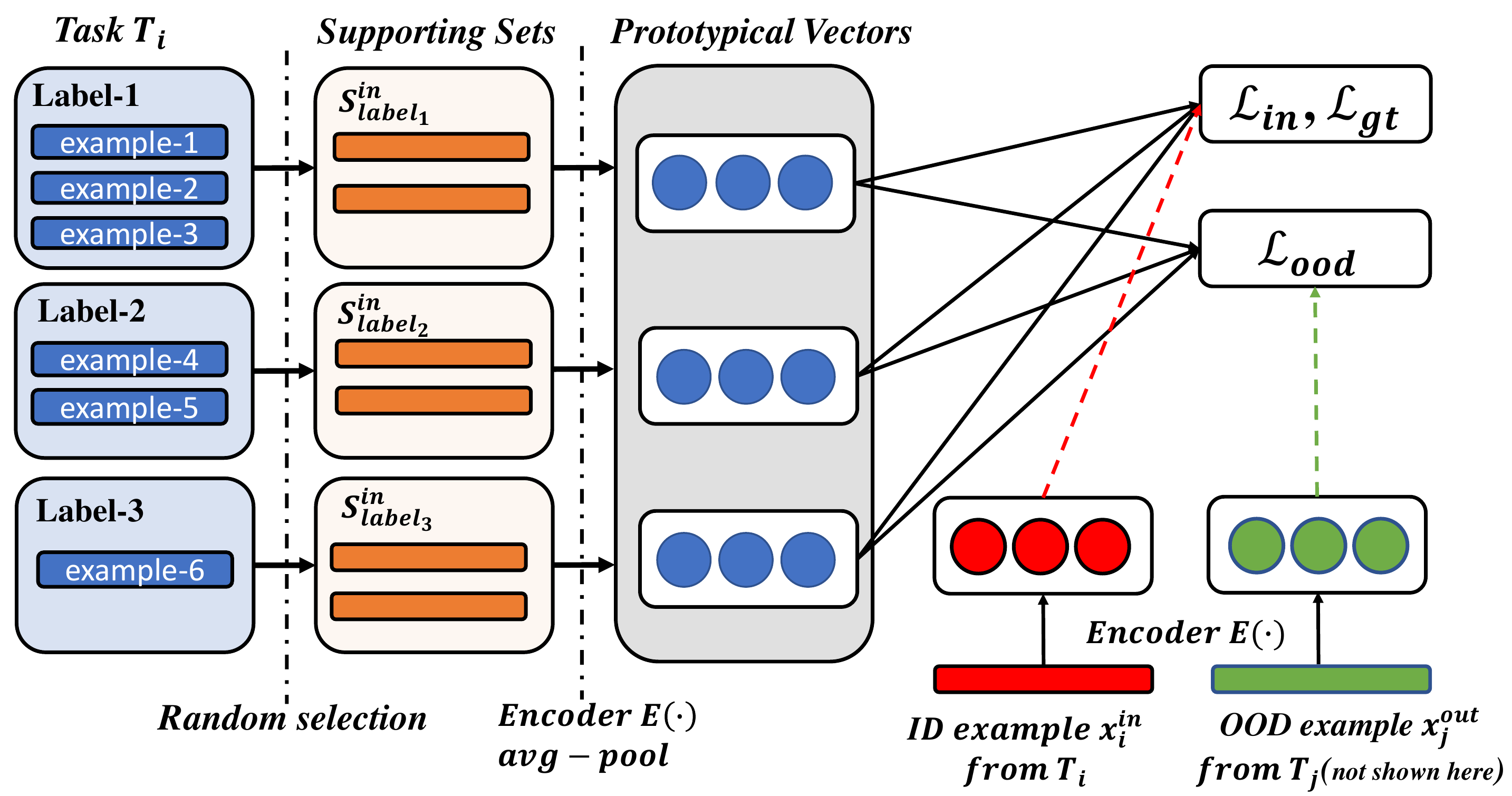}
  \caption{Model overview: the model maximizes likelihood of the ground-truth of ID example, minimizes distance between ID example and ground-truth, and maximizes the distance of OOD example and all ID labels.}
  \label{fig:model-overview}
  \vspace{-0.15in}
\end{figure}

We propose an OOD-resistant prototypical network for both OOD detection and few-shot ID classification. We follow \cite{prototypeNet} in few-shot image classification by training a prototypical network on $\mathcal{T}$ and directly perform prediction on $D$ without additional training.  But our method is different from the prior work in that during the meta-training, while we maximize the likelihood of the true label for an example in $T_i$, we also sample an example from another meta-train task $T_{j}$ for the purpose of OOD training by maximizing the distance between the OOD instance and the prototypical vector of each ID label.

\subsection{General Framework}
As in Fig. \ref{fig:model-overview}, on a large-scale source dataset $\mathcal{T}$ with the following steps: 
\vspace{-0.1in}
\begin{enumerate}
\itemsep-0.33em
  \item Sample a training task $T_i$ from $\mathcal{T}$  (e.g., the Book category of Amazon Review in Section 4), and another task $T_j$  from $\mathcal{T}-T_i$ (e.g. the Apps-for-Android category). 
  \item Sample an ID training example $x_i^{in}$ from $T_i$, and a simulated OOD example $x_j^{out}$ from $T_j$. 
  \item Sample $N$ labels ($N$=4) from $T_i$ in addition to the label of $x_i^{in}$. For the ground-truth label and $N$ negative labels, we select $K$ training examples for each label ($K$-shot learning, we set $K$=20). If a label has less than $K$ examples, we replicate the selected example to satisfy $K$. Therefore, $(N+1) \times K$  examples serve as a supporting set $\mathcal{S}^{in} = \{S_{l}^{in}\}_{l=1}^N$. 
  \item Given a batch of dynamically-constructed meta-train set ($x_i^{in}, x_j^{out}, \mathcal{S}^{in}$), $E(\cdot)$ encodes $x_i^{in}$, $x_j^{out}$ and the examples in $S_{l}^{in}$ using a deep network (Any DL structure can be used for the encoder, such as LSTM and CNN. Here we use a one-layer CNN with a mean pooling. The detailed CNN hyper-parameters are introduced in Section \ref{section_exp_results}).
  \item Following \cite{prototypeNet}, a Prototypical Vector representation for each label is generated, by averaging all the examples' representations of that label.
  \item The model is optimized by an objective function, defined by  $x_i^{in}$ ,  $x_j^{out}$  and $\mathcal{S}^{in}$. Details in Section 3.2.
  \item Repeat these steps for multiple epochs (5k in this paper) to train the model, and select the best model based on an independent meta-valid set $\mathcal{T}^{valid}$. $\mathcal{T}^{valid}$ contains tasks that are homogeneous to the meta-test task $D$.
\end{enumerate}
The only trainable parameters of this model are in the encoder $E(\cdot)$. Therefore the trained model can be easily transferred to the few-shot target domain.

\subsection{Training Objective and Runtime}
Prototypical networks \cite{prototypeNet} minimize a cross-entropy loss defined on the distance metrics between $x_i^{in}$ and the supporting sets,
\vspace{-1mm}
\begin{eqnarray}
  \small
  \mathcal{L}_{in} &=& - \log \frac{\exp \alpha F(x_i^{in}, S_{l_i}^{in})}{\sum_{l'} \exp \alpha F(x_{i}^{in}, S_{l'}^{in})}
  \label{eq:in}
\end{eqnarray}
where $l_i$ is the ground-truth label of $x_i$, $\alpha$ is a re-scaling factor. Here we define $F$ as a cosine similarity score (mapped to the range between 0 and 1)\footnote{Following \cite{prototypeNet}, we also tried squared Euclidean distance, but did not achieve better results.} between the $E(\cdot)$-encoded representations of $x$ and the prototypical vector of a label. Our experiments show this meta-learning approach is efficient for ID classification, but is not good enough for detecting the OOD examples.

We propose two more training losses in addition to the ${L}_{in}$ for OOD detection. The rationale behind this addition is to adopt the examples from other tasks as simulated OOD examples for the current meta-train tasks. Specifically, we first define a hinge loss on $x_j^{out}$ and the closest ID supporting set in $\mathcal{S}^{in}$, then we push the examples from another task away from the prototypical vectors of ID supporting sets. 
\vspace{-1mm}
\begin{eqnarray}
\small
  \mathcal{L}_{ood} = max[0, \max_{l}(F(x_j^{out}, S_l^{in}) - \mathcal{M}_1)]
  \label{eq:ood}
\end{eqnarray}
We expect optimizing only on $\mathcal{L}_{in}$ and ${L}_{ood}$ will lead to lower confidences on ID classification, because the system tends to mistakenly reduce the scale of $F$ in order to minimize the loss for OOD examples. 
Therefore we add another loss to improve the confidence of classifed ID labels. 
\vspace{-1mm}
\begin{eqnarray}
\small
\small
  \mathcal{L}_{gt} &=& max[0, \mathcal{M}_2 - F(x_i^{in}, S_{l_{i}}^{in}))]
  \label{eq:gt}
\end{eqnarray}
The model is optimized on the three losses. 
\vspace{-1mm}
\begin{eqnarray}
  \mathcal{L} &=& \mathcal{L}_{in} + \beta \mathcal{L}_{ood} + 
  \gamma \mathcal{L}_{gt}
  \label{eq:all}
\end{eqnarray}
where $\alpha$, $\beta$, $\gamma$, $\mathcal{M}_1$ and $\mathcal{M}_2$ are hyper-parameters, whose detailed values are shown in Section \ref{section_exp_results} .



During inference, the supporting set per label is generated by averaging the encoded representations of all instances of that label in $D^{train}$ and the prediction is based on $F(x, S_l^{in})$. OOD detection is decided with a \emph{confidence} threshold.

\section{Datasets}

Our methods are evaluated on two datasets and each has many tasks and is divided into meta-train, meta-valid and meta-test sets, which are respectively used for background model training, evaluation and hyper-parameter selection.

\paragraph{Amazon Review \footnote{We will release Amazon data and our code at https://github.com/SLAD-ml/few-shot-ood  }:}
We follow \cite{diverse_few_shot} to construct multiple tasks using the Amazon Review dataset \cite{he2016ups} . We convert it into a binary classification task of labeling the review sentiment (positive/negative). It has 21 categories of products, each of which is treated as a task. 
We randomly picked 13 categories as meta-train, 4 as meta-test and 4 as meta-valid. (another 3 original categories are discarded due to not enough examples to make a dataset). We construct a 2-way 100-shot problem per meta-test task by sampling 100 reviews per label in a category. For the test examples in meta-test and meta-valid, we sample other categories' examples as OOD, merged with a equal number of ID instances. We used all available data for meta-train. 


\paragraph{Conversation Dataset:}
An intent classification dataset for a AI conversational system. It has 539 categories/tasks. We allocate 497 tasks as meta-train, and 42 tasks as meta-test. This dataset is different and more difficult than the typical ID few-shot learning data: 1) Both the meta-test and meta-train tasks are not restricted to $N$-way $K$-shot classification, and the source dataset is highly imbalanced across labels; 2) Each task has a variety of labels (utterance intents), whereas Amazon data always has two labels. There are 29\% OOD testing instances in meta-test, which are human-labeled and not generated from other tasks.


\section{Experimental Results}
\label{section_exp_results}

\begin{figure*}
  \begin{minipage}[b]{0.60\textwidth}
\small
\centering

\begin{tabular}{lllllll}
\toprule
 & \multicolumn{3}{c}{\underline{Conversation}} & \multicolumn{3}{c}{\underline{Amazon Review}} \\
 (\%) & EER & CER & Comb. & EER & CER & Comb. \\
\midrule
OSVM & 63.6 & - & - & 47.6 & - & - \\
LSTM AutoEnc. & 48.0 & 78.4 & 79.5 & 45.4 & 29.3 & 38.6  \\
Vanilla CNN & 26.4 & 76.8 & 77.6 & 47.7 & 34.4 & 42.8 \\
\hline
Proto. Network & 26.9 & 32.5 & 44.5 & 46.5 & \textbf{7.3} & 47.6 \\
O-Proto ($\mathcal{L}_{in}+\mathcal{L}_{gt}$) & 27.6 & 33.3 & 46.2 & 47.8 & 7.4 & 48.9 \\
O-Proto ($\mathcal{L}_{in}+\mathcal{L}_{ood}$) & 24.5 & 30.1 & 41.2 & 24.7 & 9.7 & 30.1 \\
O-Proto (all) & \textbf{24.1} & \textbf{29.6} & \textbf{40.8} & \textbf{24.0} & 9.1 & \textbf{29.1} \\

\hline
Proto. with bilstm & 25.0 & 32.5 & 42.6 & 45.1 & \textbf{6.8} & 46.0 \\
O-Proto with bilstm & \textbf{22.0} & \textbf{30.5} & \textbf{39.8} & \textbf{21.9} & 9.0 & \textbf{27.1}  \\
\bottomrule
\end{tabular}
\captionof{table}{O-Proto is compared with other baselines for Conversation and Amazon data}
\label{tab:intent}
    \vspace{-0.22in}
  \end{minipage}
  \begin{minipage}[b]{0.37\textwidth}
    \centering
  \includegraphics[scale=0.36]{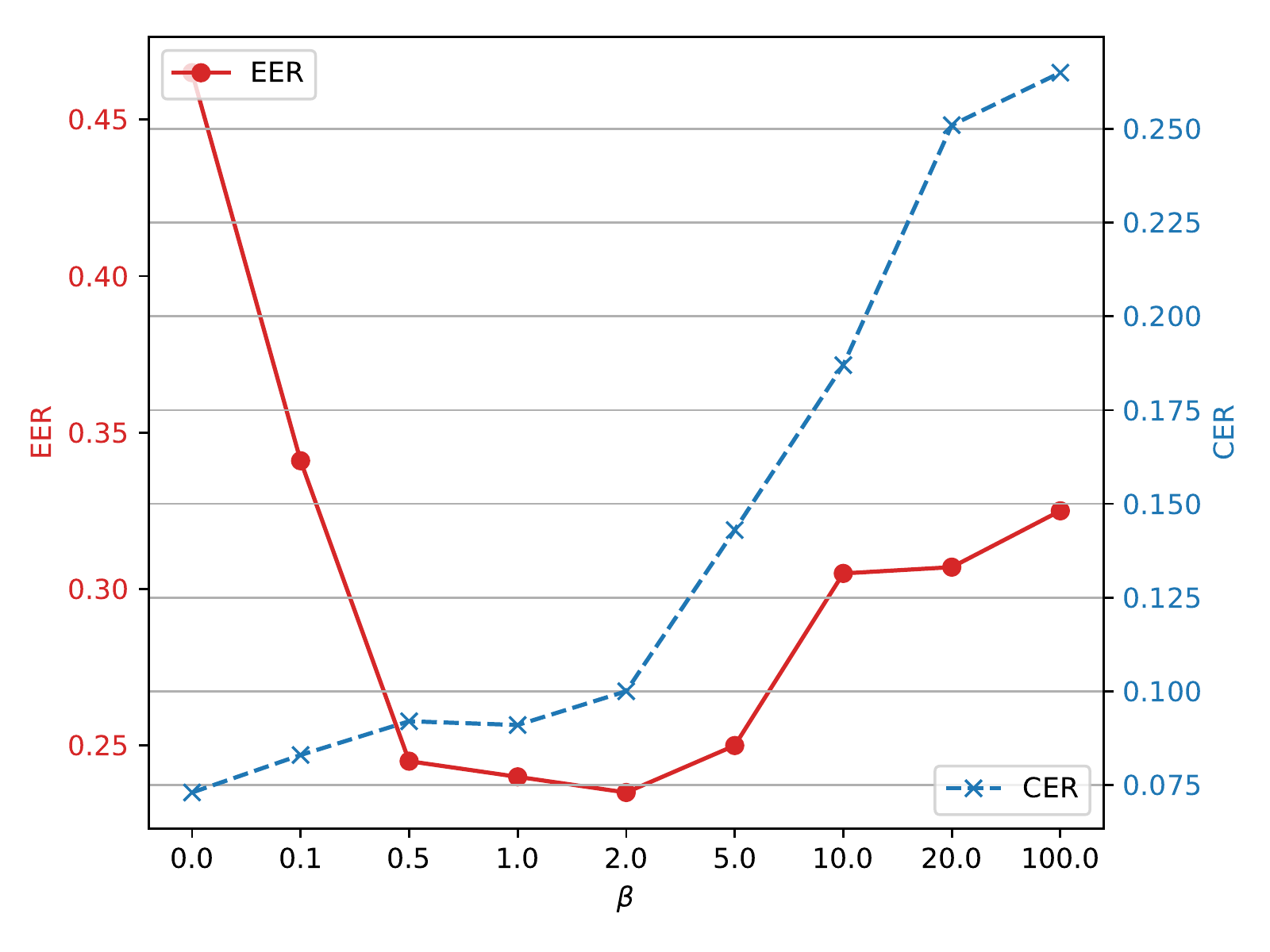}
  \caption{\small Various $\beta$ for Amazon data}
  \small
  \label{fig:beta}
   \vspace{-0.13in}
    \end{minipage}
  \end{figure*}

\noindent \textbf{Baselines:}
We compare our model \textbf{O-Proto} with 4 baselines: 1) \textbf{OSVM \cite{osvm}}: OSVM is trained on meta-test set, and learn a domain boundary by only examining ID examples. 
2) \textbf{LSTM-AutoEncoder \cite{ood_sent_detect}}: Recent work on OOD detection that uses only ID examples to train an autoencoder for OOD detection.
3) \textbf{Vanilla CNN}: A classifier with a typical CNN structure that uses a confidence threshold for OOD detection.
4) \textbf{Proto. Network \cite{prototypeNet}}: A native prototypical network trained on $\mathcal{T}$ with only the loss $\mathcal{L}_{in}$, which uses a confidence threshold for OOD detection.
We test the Proto. Network with both CNN and bidirectional LSTM as the encoder $E(\cdot)$.

\begin{figure}[t!]
  \centering
  \includegraphics[scale=0.36]{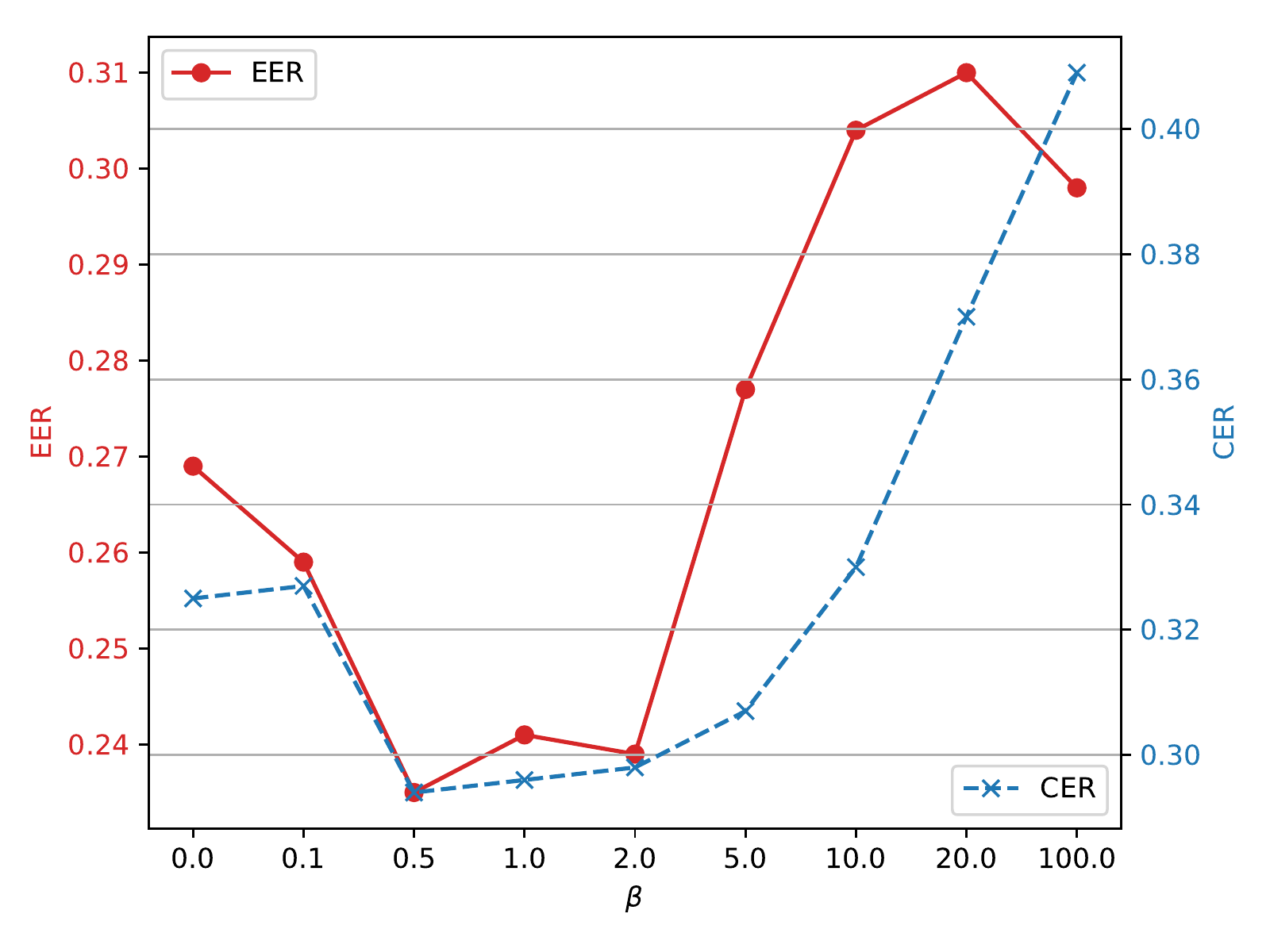}
  \caption{Various $\beta$ for Conversation data}
  \label{fig:beta-wa}
  \vspace{-0.16in}
\end{figure}

\paragraph{Hyper Parameters:} We introduce the  hyper-parameters of our model and all baselines below. 

We use Python scikit-learn One-Class SVM as the basis of our OSVM implementation. We use Radial Basis Function (RBF) as the kernel and the gamma parameter is set to auto. We use squared hinge loss and L2 regularization.

We follow the same architecture as proposed in \cite{ood_sent_detect} for the LSTM-Autoencoder. In LSTM, we set the input embedding dimension as 100 and hidden as 200. We use RMSprop as the optimizer with a learning rate of 0.001. We train the LSTM with a batch size of 32 and 100 epochs. For Autoencoder, we set the hidden size as 20. We use Adam as the optimizer with a learning rate of 0.001. We train the model with a batch size of 32 for 10 epochs.

For vanilla CNN, we use the most common CNN architecture used in NLP tasks, where the convolutional layer on top of word embedding has 128 filters followed by a ReLU and max pooling layer before the final softmax. We use Adam as the optimizer with a learning rate of 0.001. We train the model with a batch size 64 for 100 epochs.

Our proposed model O-Proto uses the similar CNN architecture, the optimizer and the learning rate in the previous Vanilla CNN. The input word embeddings are pre-trained by 1-billion-token Wikipedia corpus. We set the batch size as 10. In Eq. \ref{eq:in}, \ref{eq:ood}, \ref{eq:gt} and \ref{eq:all}, $\alpha$, $\beta$, $\gamma$, $\mathcal{M}_1$ and $\mathcal{M}_2$ are hyper-parameters, which we fix $\beta$, $\gamma$ as 1.0 by default, and set $\alpha$, $\mathcal{M}_1$ and $\mathcal{M}_2$ as 10.0, 0.4 and 0.8 according to the meta-valid performance of Amazon dataset. The sentence encoder, CNN, has 200 filters, followed by a tanh and mean pooling layer before the final regression layer. The maximum length of tokens per example is 40, and any words out of this range will be discarded. 
During training, we set the size of sampled negative labels Step 3 (section 3.1) to at most four, so there will be maximum five labels involved in a training step (1 positive, 4 negative). The supporting set size for each label are 20. 

To make a fair comparison, we follow the same hyper-parameters as O-Proto in Proto. Network, except that the weight of $\mathcal{L}_{ood}$ and $\mathcal{L}_{gt}$, $\beta$ and $\gamma$, are set to zero. 

\vspace{1mm}
\noindent \textbf{Evaluation Metrics:}
Following~\cite{ood_sent_detect,ood-utt}, we use a commonly used OOD detection metric, equal error rate (\textbf{EER}), which is the error rate when the confidence threshold is located where false acceptance rate (\textbf{FAR}) is equivalent to false rejection rate (\textbf{FRR}).

\vspace{-1.8mm}
{\small
\begin{eqnarray*}
FAR &=& \frac{Number\:of\:accepted\:OOD\:sentences}{Number\:of\:OOD\:sentences}\\
FRR &=& \frac{Number\:of\:rejected\:ID\:sentences}{Number\:of\:ID\:sentences}
\end{eqnarray*}
}%
We use class error rate \textbf{CER} to reflect ID performance.
Lastly, we applied the threshold used in the EER to ID test examples to test how many ID examples are rejected by the OOD detection, as \textbf{Combined-CER}~(Comb. in Table \ref{tab:intent}).

\vspace{0.05in}
\noindent \textbf{Results:}
Table \ref{tab:intent} compares the proposed model and baselines on EER and CER on the two datasets. \footnote{No CER reported for \textit{OSVM} as it treats ID as one class and does not support ID classification.} For prototypical-network related models, we randomly initialize the model parameters by 10 times and report the averaged metrics. \textit{OSVM, LSTM-AutoEncoder and Vanilla CNN} perform poorly on OOD detection as expected, as they require larger number of ID examples which is not a few-shot scenario. \textit{Proto.Network} has a better performance on ID classification. But it is not designed for OOD, thus it does not perform well on OOD detection. \textit{O-Proto} achieves significantly better EERs (2.8\% improvement on Conversation, and 22.5\% on Amazon), yielding competitive results on CER compared to \textit{Proto.Network}. These lead to a remarkable improvement on Combined-CER. 
\emph{O-Proto} improves less on EER in Conversation than Amazon, because some Conversation tasks actually come from the conversational service providers belonging to similar business domains. Our model is better even when meta-train datasets are from slightly different domains. Moreover, in Table \ref{tab:intent} we show the ablation study by removing $\mathcal{L}_{ood}$ and $\mathcal{L}_{gt}$ from \textit{O-Proto}, respectively. \textit{O-Proto} without $\mathcal{L}_{ood}$ completely loses the ability of OOD detection. Compared to the one without $\mathcal{L}_{gt}$, \textit{O-Proto} with all losses gives a mild improvement. We also observe a more stable testing performance among epochs during training. 
Finally, we replace the CNN encoders with bidirectional LSTMs (the bottom of Table 1), which yields the same dimension of sentence representations as CNN. For Conversation data, we achieve the best performance on validation set when $\alpha$ and $\gamma$ are 0.5. We observe comparable performances with respect to \textit{Proto.Network} and 
\emph{O-Proto}, showing that our proposed OOD approach is not limited to a specific sentence encoding architecture.


\vspace{0.05in}
\noindent \textbf{Improvement in different K-shot settings:}
On the Amazon data, we construct different $K$-shot tasks as meta-test (results shown in Table \ref{tab:ablation}), and observe consistent improvements on EER.

\begin{table}
\small
\centering
\setlength\tabcolsep{4pt}
\begin{tabular}{l|ccccc}
\toprule
 EER (\%) & K=1 & =5 & =10 & =20 & =100\\
\midrule
\begin{tabular}[c]{@{}l@{}}Proto. \end{tabular} & 53.9 & 46.4 & 44.5 & 41.0 & 46.5 \\
\begin{tabular}[c]{@{}l@{}}O-Proto. \end{tabular} & 31.1 & 25.1  & 24.0 & 23.4 & 24.1 \\
\bottomrule
\end{tabular}
\vspace{-2mm}
\caption{Compare our model with Prototype Network on EER when choosing various $K$-shot values}
\label{tab:ablation}
\vspace{-4mm}
\end{table}


\vspace{0.05in}
\noindent \textbf{Effect of $\beta$:}
Fig. \ref{fig:beta} and \ref{fig:beta-wa} show EER and CER with different $\beta$ values on the two datasets. We observe within a proper range of $\beta$ (between 0.5 and 2.0), the model can provide stable improvement on EER, guaranteeing competitive CER results. 
\section{Conclusion}
\vspace{-2mm}
Inspired by the Prototypical Network, we propose a new method to tackle the OOD detection task in low-resource settings. 
Evaluation on real-world datasets demonstrates that our method performs favorably against state-of-the-art algorithms on the OOD detection, without adversely affecting performance on the few-shot ID classification.


\bibliography{emnlp-ijcnlp-2019}
\bibliographystyle{acl_natbib}

\newpage

\clearpage

\end{document}